%% file: main.tex
\documentclass[runningheads]{llncs}

\usepackage{eccv}

\usepackage{eccvabbrv}

\usepackage{graphicx}
\usepackage{booktabs}

\usepackage[accsupp]{axessibility}  

\usepackage{hyperref}

\usepackage{orcidlink}

\usepackage{wrapfig}

\usepackage{tabularx}

\usepackage{multicol}
\usepackage{multirow}

\usepackage{booktabs}
\usepackage{makecell} 
\usepackage{array} 
\usepackage{relsize} 
\newcommand{\medium}{\fontsize{8.3}{12}\selectfont}
\newcommand{\qhrh}{\fontsize{6.5}{12}\selectfont}

\usepackage{amssymb}
\usepackage{pifont}

\definecolor{cb_orange}{rgb}{1.0,0.51,0.0}
\definecolor{cb_blue}{rgb}{0.22,0.49,0.72}
\definecolor{cb_green}{rgb}{0.3,0.67,0.29}
\definecolor{cb_red}{rgb}{0.89,0.1,0.11}
\definecolor{cb_pink}{rgb}{1, 0, 0.4}

\begin{document}


\title{Co-synthesis of Histopathology Nuclei Image-Label Pairs using a Context-Conditioned Joint Diffusion Model}

\titlerunning{Co-synthesis of Histopathology Nuclei Image-Label Pairs}


\author{Seonghui Min$^*$ \and
Hyun-Jic Oh$^*$ \and
Won-Ki Jeong$^\dagger$}

\def\thefootnote{*}\footnotetext{Equal contribution}
\def\thefootnote{$\dagger$}\footnotetext{Corresponding author: wkjeong@korea.ac.kr}

\authorrunning{S.~Min et al.}

\institute{
College of Informatics, Department of Computer Science and Engineering, \\
Korea University, Seoul, South Korea
\\
}


\maketitle

\input{00_abstract}

\input{01_introduction}
\input{02_related_work}

\begin{figure}[tb!]
    \centering
    \includegraphics[width=\textwidth, keepaspectratio]{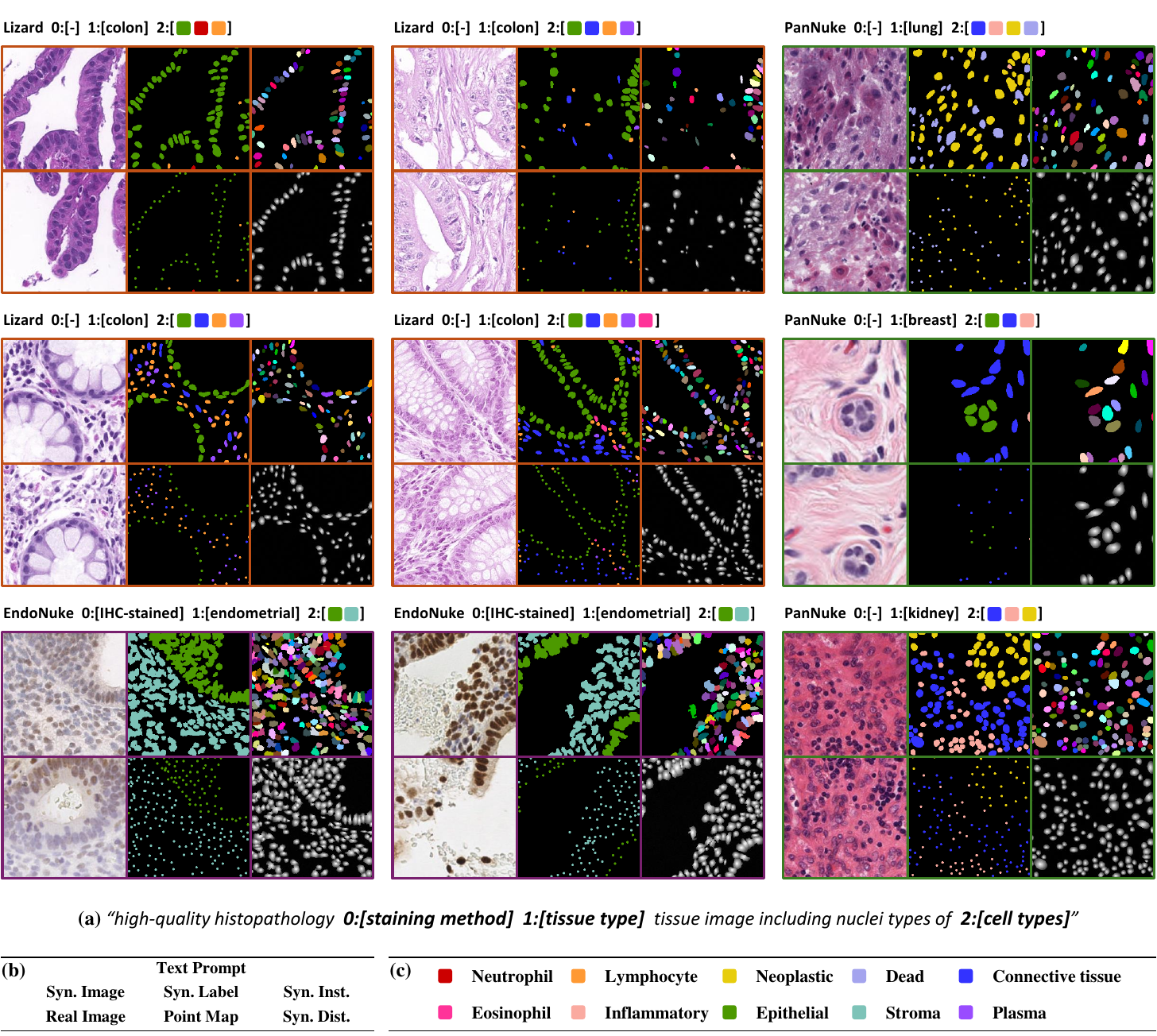} 
    \caption{
    Synthetic samples on the Lizard~\cite{lizard}, PanNuke~\cite{pannuke}, and EndoNuke~\cite{endonuke} datasets generated by our method.
    Each set of paired synthetic images, semantic labels, instance masks, and distance maps is shown with conditional point maps and text prompts.
    Real images are also included for color comparison with the corresponding synthetic images.
    (a) is a frame for the text prompting.
    (b) indicates the arrangement of each component.
    (c) provides a color legend for the classes of nuclei.
    }
    \label{fig:main_quali}
\end{figure}

\input{03_methods}

\input{04_experiments}

\input{05_conclusion}

\subsubsection{Acknowledgements} 
This work was partially supported by the National Research Foundation of Korea (RS-2024-00349697, NRF-2021R1A6A1A13044830), the Institute for Information \& Communications Technology Planning \& Evaluation (IITP-2024-2020-0-01819), the Technology development Program(RS-2024-00437796) funded by the Ministry of SMEs and Startups(MSS, Korea), and a Korea University Grant.

\clearpage
\input{06_supplementary}
\clearpage

%
%

\bibliographystyle{splncs04}
\bibliography{main}
\end{document}

%% file: 00_abstract.tex
\begin{abstract}


In multi-class histopathology nuclei analysis tasks, the lack of training data becomes a main bottleneck for the performance of learning-based methods.
%
To tackle this challenge, previous methods have utilized generative models to increase data by generating synthetic samples.
However, existing methods often overlook the importance of considering the context of biological tissues (e.g., shape, spatial layout, and tissue type) 
in the synthetic data. 
Moreover, while generative models have shown superior performance in synthesizing realistic histopathology images, none of the existing methods are capable of producing image-label pairs at the same time. 
%
%
In this paper, we introduce a novel framework for co-synthesizing histopathology nuclei images and paired semantic labels 
using a context-conditioned joint diffusion model.
%
We propose 
conditioning of a diffusion model using nucleus centroid layouts 
with structure-related text prompts
to 
incorporate spatial and structural context information into the generation targets.
%
Moreover, we enhance the granularity of our synthesized semantic labels by generating instance-wise nuclei labels using distance maps synthesized concurrently in conjunction with the images and semantic labels. 
%
We demonstrate the effectiveness of our framework in generating high-quality samples on multi-institutional, multi-organ, and multi-modality datasets.
Our synthetic data consistently outperforms existing augmentation methods in the downstream tasks of nuclei segmentation and classification.

\keywords{Joint diffusion model \and Data augmentation \and Histopathology nuclei segmentation}

\end{abstract}

%% file: 01_introduction.tex
\section{Introduction}
\label{sec:introduction}

%
Cell nuclei segmentation and classification are crucial tasks in digital pathology for examining nuclear characteristics, such as size, shape, density, etc, which provide important evidence for disease diagnosis~\cite{cheng2020computational}.
Given the complicated nature of histopathology images, the adoption of deep learning-based computer-aided diagnosis has now become the de facto standard in computational pathology~\cite{verghese2023computational}. 
%
%
Indeed, numerous fully-supervised learning approaches have been introduced and proven to be effective in nuclei analysis~\cite{graham2019hover, doan2022sonnet, transnuseg}. 
However, the potential of learning algorithms is limited by the lack of appropriate training data. 
This is primarily attributed to the labor-intensive nature of manually generating annotations, necessitating the domain expertise of pathologists for accurate labeling. 
%

\begin{figure}[t!]
    \centering
    \includegraphics[width=\textwidth, keepaspectratio]{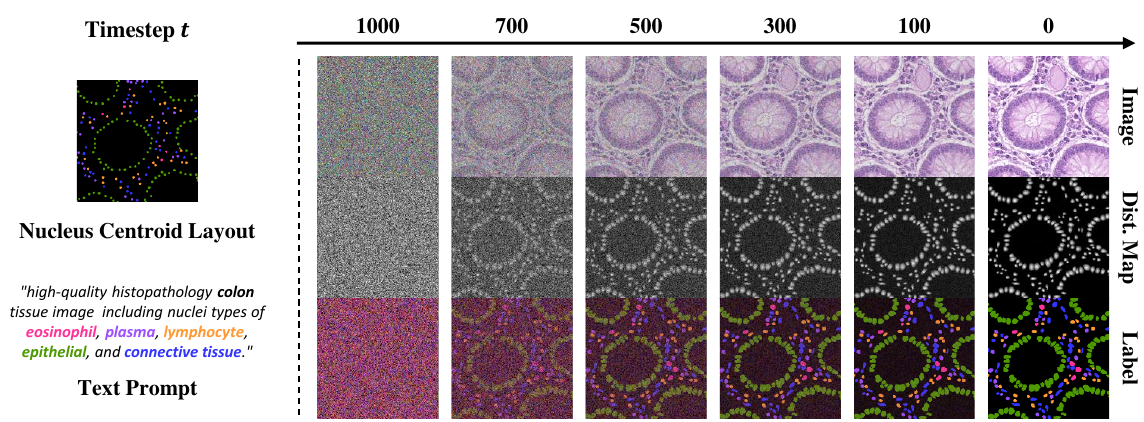} 
    \caption{Given a nucleus centroid layout and structure-related text prompt, our proposed approach generates a pair of histopathology nuclei image, distance map, and semantic label that aligns with the specified conditions. This process is concurrently performed by a single joint diffusion model.}
    \label{fig:teaser}
\end{figure}

One notable research direction to address data insufficiency is synthetically generating the training data (i.e., data augmentation). 
Recently, there has been a growing interest in using generative models based on GANs~\cite{gan} and diffusion models~\cite{dpm} for data synthesis. 
Specifically, diffusion models have demonstrated superior performance compared to GANs in both natural and histopathology image synthesis~\cite{diffusion_beats_gan, diffuse_morph}. 
Furthermore, the incorporation of semantic label conditional diffusion models~\cite{sdm} has gained prominence for generating images that seamlessly align with provided labels.
%
%
%
In histopathology nuclei image synthesis, the majority of current research relies on pixel-level segmentation labels to generate synthetic histopathology images that are accurately aligned with the provided labels.
Some studies~\cite{sian, nasdm} employed the original reference labels as is, while others introduced random perturbations to the original labels~\cite{gong2021style, diffmix} or generated randomly distributed labels prior to image generation~\cite{hou2019robust, sharp-gan, diffaug}. 
%
%
Although these methods are effective in generating histopathology image-label pairs, they may not be as versatile in producing diverse labels and do not faithfully reconstruct spatial and structural contexts in real histopathology specimens. 

The primary motivation driving our work arises from the observation of a critical gap in existing methods - none are capable of producing image-label pairs with user-controllable image content (e.g., tissue type) 
in a highly reliable spatial context. 
For example, Semantic-Palette~\cite{semantic-palette} allows for the manipulation of class proportions during layout generation, but it cannot precisely control the spatial placement of labels. 
On the other hand, Abousamra~\textit{et al.}~\cite{abousamra2023topology} introduced a technique for generating cell layouts alongside corresponding histopathology images, adhering to specified spatial contexts. 
However, its applicability is limited to detection and classification tasks due to the inherent nature of point labels.
Therefore, our idea is to bridge the gap between these two methodologies by incorporating a point map condition representing nucleus centroids and a text condition representing tissue and nuclei types in the generation of histopathology images and their corresponding multi-class nuclei segmentation labels, empowering users with complete control over the spatial layout and content of cell images. 
%
Additionally, we also observed a common issue in conventional nuclei label synthesis - nuclei instances in the resulting semantic labels tend to be closely located and clustered into a larger one.
%
%
To address this issue, we propose the generation of distance maps alongside images and labels, which can be directly used to separate individual instances. 
%
%
The overview of the proposed method is shown in~\cref{fig:overview}.
The main contributions of our work are several-fold as follows:

\begin{figure}[t!]
    \centering
    \includegraphics[width=\textwidth, keepaspectratio]{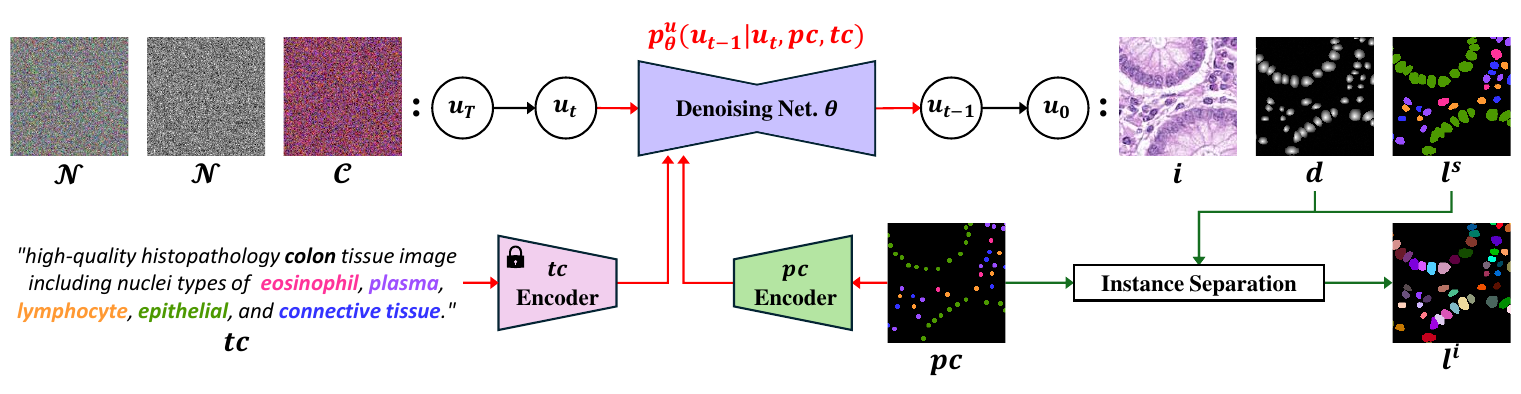}
    \caption{Overview of the proposed method. We formulate joint diffusion process to synthesize multiple targets: image $i$, distance map $d$, and semantic label $l^s$.
    We utilize conditioning of text $tc$ and point map $pc$ to improve the sample quality and provide controllable capacity. We generate highly accurate instance label $l^i$ using $pc$, $d$, and $l^s$.
    }
    \label{fig:overview}
\end{figure}

\begin{itemize}
    \item 
    We propose a co-synthesis framework for multi-class nuclei datasets that generates images with semantic and instance labels using a single 
    diffusion model. 
    The proposed method models the joint distribution of histopathology images, semantic labels, and distance maps, enabling the simultaneous generation of the whole targets.
%
%
    \item 
    We incorporate two conditions, a nucleus centroid layout and a text prompt, to enhance the model's capacity to capture the critical and intricate visual context of 
    real histopathology specimens.
    This not only gives the user full control over the tissue and nuclei types 
    but also provides flexibility in the design of the spatial arrangement.
%
%
    \item 
        We generate high-quality instance labels by separating individual nuclei in the semantic labels using distance maps and the conditioned nucleus centroid layouts. 
        The generated instance labels are necessary for high-level nuclei analysis such as state-of-the-art nuclei segmentation algorithms.
%
%
    \item 
    We demonstrate the efficacy of our approach using multi-institutional, multi-organ, and multi-modality datasets through quantitative and qualitative assessment of downstream tasks, including nuclei segmentation and classification. 
\end{itemize}

%% file: 02_related_work.tex
\section{Related Work}
\label{sec:related_work}


\subsection{Generative Models for Image-Label Synthesis}

The task of generating image-label pairs aims to model the joint distribution between images and their associated labels.
This research area has seen significant developments, greatly influenced by generative adversarial neural networks (GANs)~\cite{gan}. For instance, Dataset-GAN~\cite{dataset-gan} initially generates images and then uses GAN latent codes to generate corresponding semantic labels. 
Alternatively, SB-GAN~\cite{sb_gan} reverses this process: starting with label generation and then synthesizing images based on these generated labels. 
Semantic-Palette~\cite{semantic-palette} introduces controllable class proportions in the process of generating semantic layouts.
%

A recent milestone in generative models has been the emergence of the Denoising Diffusion Probabilistic Model (DDPM)~\cite{dpm}. 
DDPM stochastically simulates the denoising process and has showcased superior performance over state-of-the-art generative models~\cite{diffusion_beats_gan}. 
Notably, the diffusion model has consistently outperformed traditional GAN-based approaches in various studies~\cite{dataset-ddpm, dataset-diffusion} on generating image-label pairs.
%
Recently, 
Park~\textit{et al.}~\cite{gcdp} have introduced an approach for co-synthesizing image-label pairs using a single diffusion model for the text-to-image synthesis task. 
Their method efficiently captures the joint distribution of image-label pairs by applying a Gaussian diffusion process to the images and a categorical diffusion process to the labels.

\subsection{Histopathology Image Synthesis}


Extensive research has been dedicated to this field, with a focus on leveraging generative models for data synthesis. 
Notably, recent diffusion model-based approaches have exhibited remarkable superiority over GAN-based methods~\cite{attribute_gan, diffuse_morph, nasdm}.
%
%
%

For nuclei image synthesis, 
previous research has often used original labels as a reference due to the requirement of domain-specific knowledge to annotate histopathology images. 
For instance, SIAN~\cite{sian} employed the original reference labels as conditioning, yielding a diverse set of stylized images using encoded style vectors for a multi-organ, single-class nuclei dataset. 
NASDM~\cite{nasdm}, on the other hand, synthesized images by referencing the original labels, leveraging a semantic label conditional diffusion model (SDM)~\cite{sdm}. 
Alternatively, 
InsMix~\cite{insmix} and DiffMix~\cite{diffmix} applied random label perturbations, such as copying and pasting or random adjustments to nuclei positions, before generating images based on these modified labels. 
In addition, some methods~\cite{hou2019robust, sharp-gan, diffaug} generated randomly distributed labels and used them as the basis for image generation. 
For example, Yu~\textit{et al.}~\cite{diffaug} introduced a two-stage diffusion model framework, comprising an unconditional training approach utilizing a diffusion model for label generation followed by SDM.
%
However, this approach was designed for single-class nuclei data, lacked fine control over the label synthesis step, and involved inference from two diffusion models, resulting in prolonged data synthesis times. 
Moreover, these approaches do not deeply consider the spatial details in the original labels when altering or generating new ones. 
In contrast, Abousamra~\textit{et al.}~\cite{abousamra2023topology} introduced a method focused on generating structure-aware point layouts, emphasizing the significance of capturing spatial and structural correlations related to nuclei positions. Nevertheless, this approach generated point labels rather than complete pixel-level labels.

In this paper, we introduce 
a unified framework to generate histopathology images and their corresponding labels simultaneously using a single joint diffusion model.
%
Furthermore, we introduce nucleus centroid layout and text conditioning for better control over nuclei positioning and enhance the structural realism of the synthesized pairs.
Lastly, we incorporate highly accurate nuclei instance labels through post-processing, expanding the applicability of the generated dataset to high-level nuclei analysis.

%% file: 03_methods.tex
\section{Methods}
\label{sec:methods}

\subsection{Background: Diffusion Models}

Denoising diffusion probabilistic models convert noise with a specific simple distribution into data sampled from a more intricate distribution~\cite{dpm}.
Diffusion models employ a noise schedule 
denoted as $\beta$ during the forward process to add noise to the actual data. 
Sequential denoising then occurs in the reverse process across time steps $t \in [1, 2, ..., T]$, resulting in the generation of synthetic data $x_0$ from the noise $x_T$. 
Various noise distributions have been systematically investigated to align with the characteristics of the target data.

The Gaussian diffusion model is commonly used 
for synthesizing continuous distribution data such as images. 
The forward process with Gaussian noise that follows a normal distribution can be described as:
\begin{align}
\label{eq:guas_for}
   q(x_t\,|\,x_{t-1}) = \mathcal{N}(x_t;\, \sqrt{1 - \beta_t}x_{t-1},\, \beta_t I).
\end{align}

The categorical diffusion model proposed by Hoogeboom~\textit{et al.}~\cite{hoogeboom} is designed to synthesize discrete distribution data such as texts and segmentation labels. 
For $K$ categories, using the categorical distribution $\mathcal{C}$, the forward process is defined as:
\begin{align}
\label{eq:cate_for}
   q(x_t\,|\,x_{t-1}) = \mathcal{C}(x_t;\, (1 - \beta_t)x_{t-1} + \beta_t/K).
\end{align}

The reverse process $p_{\theta}(x_{t-1}\,|\,x_{t})$ unfolds with a deep neural network $\epsilon_{\theta}$.
Different types of diffusion models primarily focus on learning the denoising step transitioning from $t$ to $t-1$.
Consequently, the definition of the training loss is formulated as follows:
\begin{equation}
    \mathcal{L} = \mathbb{E}_{t,x_0,\epsilon} \left[ \left\| \epsilon - \epsilon_\theta(x_t, t) \right\|^2_2 \right],
\end{equation}
where the objective is to minimize the discrepancy between the predicted noise $\epsilon_{\theta}(x_t,t)$ and the real noise $\epsilon$ in $x_t$.

\subsection{Joint Diffusion Process}

The integrative deployment of appropriate distributions facilitates the concurrent generation of multiple targets in different modalities. 
We simultaneously generate images, multi-class semantic labels, and distance maps to provide a dataset with high utility for histopathology nuclei image analysis. 
%
Distance maps are used to separate indistinguishable nuclei in semantic labels, providing specific labels for each instance. 
Further details on leveraging distance maps for instance separation are described in~\cref{subsec:isnt_sep}.

Let us denote the image, distance map, and semantic label by $i$, $d$, and $l^s$, respectively. These multi-modal elements collectively form a tripartite data unit $u:=(i,d,l^s)$. Considering the properties of each modality, we model the continuous variables $i$ and $d$ with Gaussian distributions, and the discrete variable $l^s$ with a categorical distribution. 
Subsequently, we define the reverse process for $u$, where each component undergoes an independent forward process (\cref{eq:guas_for} and \cref{eq:cate_for}), as follows:
\begin{align}
   p_{\theta}^u(u_{t-1}\,|\,u_{t})
   = p_{\theta}^i(i_{t-1}\,|\,u_{t}) \cdot p_{\theta}^d(d_{t-1}\,|\,u_{t}) \cdot p_{\theta}^{l^s}({l^s}_{t-1}\,|\,u_{t}).
\end{align}
To train the joint diffusion model, we utilize a composite objective function, defined as:
\begin{align}
   \mathcal{L}_{total} =  \lambda_i\cdot\mathcal{L}_i +  \lambda_d\cdot\mathcal{L}_d +  \lambda_{l^s}\cdot\mathcal{L}_{l^s},
\end{align}
where $\lambda_i$, $\lambda_d$, and $\lambda_{l^s}$ are weighting factors that balance the contribution of each generation target to the overall training objective.

\subsection{Context Conditions: Nucleus Centroid Layout and Text Prompt}
\begin{wrapfigure}{r}{0.49\textwidth}
\centering
\vspace{-22pt}
  \includegraphics[width=0.47\textwidth]{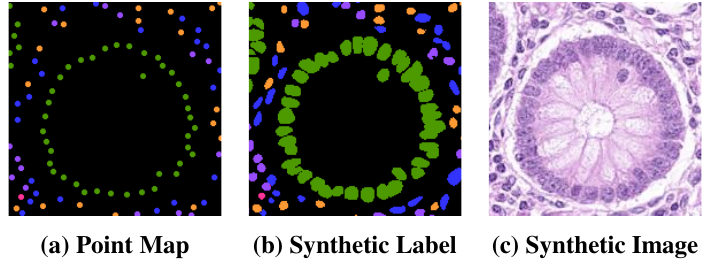} 
  \caption{Example of a point map conditioned synthetic image-label pair.
        Glands and lumens similar to those observed in real histopathology images were generated for epithelial cell points arranged in circular patterns.
        }
\vspace{-17pt}
\label{fig:pnt_exp}
\end{wrapfigure}
To generate highly realistic images and precisely control the generation process, we incorporate two types of nuclei image context conditions: nucleus centroid layout (in the form of a point map) and structure-related text prompts.

\begin{figure}[b!]
    \centering
    \includegraphics[width=\textwidth, keepaspectratio]{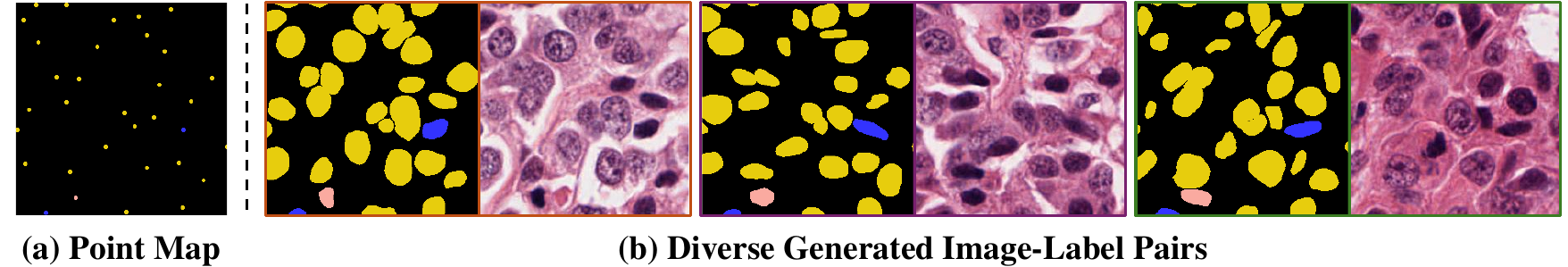} 
    \caption{
    Examples of diverse image-label pairs generated from a same point layout condition.
    }
    \label{fig:div_layout}
\end{figure}

Point map condition, indicated by $pc$, defines the centroids of the nuclei instances, providing information on their spatial positioning and class distribution, as illustrated in~\cref{fig:pnt_exp}.
This contextual overview is vital for the precise delineation and comprehension of the complex nuclear patterns found in histopathology specimens.
Moreover, $pc$, in contrast to pixel-wise constraints, provides flexibility of label generation. 
While full-label conditioned image synthesis can only diversify the generated images, our approach can generate a variety of images and labels, as shown in~\cref{fig:div_layout}.
In terms of steerability, $pc$ enables customizable data generation, allowing control over the type, quantity, and spatial configuration of nuclei.
The encoding of $pc$ is achieved through RRDBs proposed by~\cite{rrdb}.

Text condition, represented as $tc$, includes information on the tissue type of the synthetic sample and the categories of nuclei it encompasses. 
Further details on generating text prompts are described in~\cref{sec:text_gen}. We utilize PLIP~\cite{pathldm2024} to encode $tc$, which is a vision-language foundation model specialized for pathology images.

With these conditioning schemes, the output of the model is redefined as $\epsilon_\theta(u_t, t, pc, tc)$. During the
sampling,
we employ classifier-free guidance~\cite{diffusion_beats_gan} to adjust the predictive noise as follows:
\begin{align}
   \tilde{\epsilon}_\theta(u_t, t, pc, tc)
   = \omega\epsilon_\theta(u_t, t, pc, tc) + (1 - \omega)\epsilon_\theta(u_t, t, pc),
\end{align}
where $\omega$ represents the guidance scale for $tc$, balancing the trade-off between overfitting and semantic alignment.

\subsection{Nuclei Instance Separation}
\label{subsec:isnt_sep}

In this section, we delineate the methodology employed to derive instance labels, $l^i$, from $l^s$, utilizing $d$ and $pc$.
As depicted in~\cref{fig:overview}, $d$ quantifies the Euclidean distance from the centroid of each nucleus, normalized to a scale from 0 to 1. 
To separate $l^s$ at the instance level, we apply the marker-controlled watershed algorithm, as referenced in~\cite{watershed}, to $d$ and $l^s$ with $pc$ as a marker map. 
This method allows for identifying adjacent yet separate structures within the histological samples.
The integration of the synthesized $i$, $l^s$, and $l^i$ results in a data structure that encompasses not only semantic information but also instance-specific details. 
This comprehensive dataset is hence suited for a wide array of downstream applications, including segmentation and classification tasks, thus enhancing the utility of the generated synthetic histopathology images.

%% file: 04_experiments.tex
\section{Experiments}
\label{sec:experiments}

\noindent\textbf{Datasets.}
We tested 
our method
on three multi-class histopathology nuclei segmentation datasets: Lizard~\cite{lizard}, PanNuke~\cite{pannuke}, and EndoNuke~\cite{endonuke}. 
Each dataset comprises image regions derived from histopathology slides. 

Lizard is a large multi-institutional collection of six different datasets, containing colon tissue samples with 495,179 nuclei categorized into six classes.
We applied Vahadane stain normalization~\cite{vahadane} to standardize the color distribution to that of a reference slide.
We cropped patches from image regions to a size of 256$\times$256 pixels, using a stride of 128 pixels as done by NASDM~\cite{nasdm}.
We prepared 13,064 patches for the experiment with a distribution of 95\% for the training set and 5\% for the test set.

PanNuke is a multi-organ H\&E-stained dataset consisting of 19 different tissue types with 189,744 nuclei categorized into five classes.
We have omitted color normalization to preserve the color variations between the different tissue types.
The dataset consists of 7,901 patches and we divided it into 80\% of the patches for training and the remaining 20\% for testing.

\input{tables/methods_v1} 
\input{tables/fid_is_fsd}

EndoNuke is an IHC-stained dataset comprising 245,120 nuclei in endometrial tissue samples, categorized into three classes.
This dataset is designed primarily for nuclei detection tasks, but it also provides coarse semantic labels, automatically generated by watershed~\cite{wtsd} algorithm.
We selected this dataset to evaluate the applicability of our method across multiple image modalities, as it uses a different staining technique than other datasets. 
We divided a total of 1,780 patches, with a distribution of 85\% for the training set and 15\% for the test set.
For the PanNuke and EndoNuke, we used provided patches of 256$\times$256 pixels.
\\

\noindent\textbf{Text Prompt Generation.}
\label{sec:text_gen}
Since none of the datasets provide text descriptions for each sample, we generated our own prompts for text conditioning.
The text prompts include information on the tissue type and the cell types it contains, as follows:
``high-quality histopathology [tissue type] tissue image including nuclei types of [list of cell types].''
Specifically for EndoNuke, we added information on the staining method, which differs from the most commonly used H\&E staining as:
``high-quality histopathology IHC-stained [tissue type] tissue image including nuclei types of [list of cell types].''
\\

\noindent\textbf{Implementation Details.}
We used the Adam optimizer with $\beta_1 = 0.9$ and $\beta_2 = 0.99$ for model training. The learning rates were set at $10^{-4}$ for Lizard and PanNuke, and at $10^{-5}$ for EndoNuke. The weighting factors $\lambda_i$, $\lambda_d$, and $\lambda_{l^s}$ were set to 9, 1, and 3, respectively, across all datasets. 
The training batch size was 16, and we employed three separate cosine schedules, one for each output type: image, distance map, and semantic label.
The sampling step $T$ was set to 1000.
The guidance scale for the text condition was set at 3 for Lizard, 2 for PanNuke, and 0.5 for EndoNuke. All experiments were conducted using NVIDIA RTX A6000 GPUs.

\begin{figure}[tb!]
  \centering
  \includegraphics[width=\textwidth]{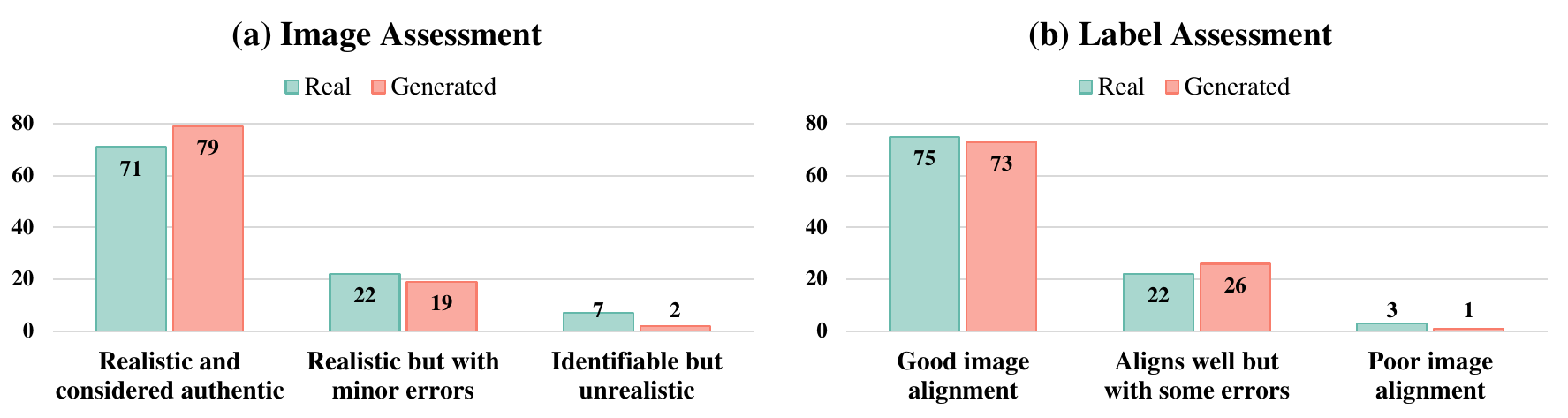}
  \caption{Pathologists' evaluation of authenticity and alignment for real image-label pairs and those generated by our method.}
  \label{fig:eval_path}
\end{figure}

\subsection{Quantitative Evaluation}


We compare the quality of generated image-label pairs quantitatively with other methods.
Since our approach is not directly comparable to existing work, i.e., no other work uses point conditions and generates image-label pairs, we compare the results with the methods using various conditioning techniques. 
%
%
Table~\ref{table:methods} shows the condition inputs required for each method and the output they generate.
We employ three metrics for the quality assessment, \text{Fr\'{e}chet} Inception Distance (FID)~\cite{FID} and Inception Score (IS)~\cite{IS} for the image, and \text{Fr\'{e}chet} Segmentation Distance (FSD)~\cite{FSD} for the label.

The comparative experiment was conducted for Yu~\textit{et al.}~\cite{diffaug}, SemanticPalette~\cite{semantic-palette}, Park~\textit{et al.}~\cite{gcdp}, SDM~\cite{sdm}, and ours, as shown in 
Table~\ref{table:fid_is_fsd}. 
The SDM composite data is guided pixel-wise from full semantic labels, resulting in high correspondence with real data in the Lizard dataset.
%
%
The instance edges generated based on instance maps are also utilized as conditions, contributing to good alignment between images and labels and enabling the generation of high-resolution images. 
Furthermore, since only images are the sole generation target, this approach focuses entirely on generating compliant quality images. 
However, SDM produced unrealistic colored images with high IS scores and poor FID scores on PanNuke, a dataset with a wide color distribution, and EndoNuke, a dataset with relatively little training data. 
Yu~\textit{et al.} generates images, labels, and distance maps without any conditions. 
Despite the similarity to the SDM in the image generation process, since Yu~\textit{et al.} was originally proposed for a single class, the quality of both images and labels deteriorated when extended to a multi-class task. 
We did not proceed with the image generation step because Yu~\textit{et al.} produces noisy and unrealistic labels. 
%
%
SemanticPalette is a method that conditions on the pixel proportion of different classes in the label. 
It has achieved commendable FSD scores because its conditioning approach and the measurement principle of FSD are similar.

Our approach involves generating image-label pairs simultaneously with a single model, incorporating appropriate noise design. 
We have achieved superior performance compared to other methods for generating image-label pairs. 
Particularly, through point conditioning, we demonstrated remarkable FSD with guidance of only 1 pixel per instance. 
Our approach demonstrated superior performance even compared to SDM, which uses full pixel labels. 
%
Even though ours achieved second place in some cases, ours scored better overall compared to other methods. 

\input{tables/mdice_v2} 

\begin{figure}[tb!]
    \centering
    \includegraphics[width=0.99\textwidth, keepaspectratio]{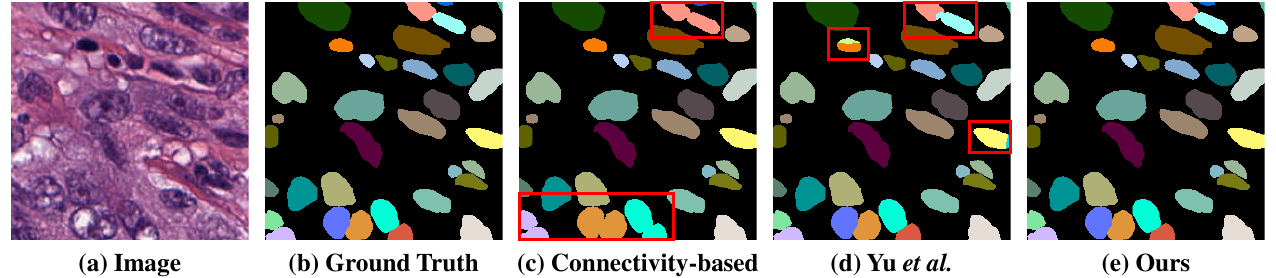}
    \caption{Comparison of instance separation algorithms.
    Under-, over-, or mis-separated instances are shown in red boxes.
    }
    \label{fig:inst_sep}
\end{figure}

\subsection{Qualitative Assessment by Pathologists}
%
As illustrated in~\cref{fig:main_quali}, our method effectively generates samples that satisfy the given conditions.
The synthetic images exhibit colors that closely resemble those of real images, and the synthetic labels are well-aligned with the paired images.
To further validate our method, we conducted an expert analysis involving the evaluation of both synthetic and real image-label pairs by five pathologists.
%
Our study included 20 synthetic images and 20 authentic pairs, focusing on evaluating their realism and alignment with their corresponding labels. 
The pathologists evaluated the images based on their expertise in pathology, including determining correspondence to actual histological structures, diagnostic quality of the tissue, color accuracy, and presence of artifacts, etc.
When evaluating the labels, they considered various criteria, such as the accuracy of cell location and type, as well as the distinction between difficult-to-identify cell types.
As shown in Fig.~\ref{fig:eval_path}a, the synthetic images received higher realism ratings than the real images. 
Meanwhile, Fig.~\ref{fig:eval_path}b shows that our synthetic data achieved image-label alignment ratings comparable to the real data. 
In addition, the pathologists rated the quality of the synthetic data pairs as good overall. 
This finding highlights the potential of generative models for histopathology data augmentation.


\subsection{Comparison of Instance Separation Methods}
We evaluate the efficiency of the point condition for instance separation using the average of the Dice coefficient (mDice)~\cite{nishimura2019weakly}, which calculates Dice score per nucleus.
Table~\ref{table:mdice} shows the results compared to traditional connectivity-based algorithms and the distance map-based watershed approach of Yu~\textit{et al.}.
Our point condition-based instance separation performed best on all datasets.
The other methods struggled to split the nuclei clusters, resulting in under-, over-, or incorrect separation as shown in Fig.~\ref{fig:inst_sep}.
Connectivity-based separation failed to separate nuclei clusters as shown in Fig.~\ref{fig:inst_sep}c. 
Distance map-based watershed algorithm worked well, but it tends to over-separate instances as shown in Fig.~\ref{fig:inst_sep}d. 


\subsection{Downstream Tasks}
We evaluated the effectiveness of synthetic data for data augmentation by assessing performance on downstream tasks such as nuclei segmentation and classification.
For training the downstream models, we used the image patches not utilized in training the diffusion model. 
%
We synthesized image-label pairs based on the point layout extracted from these patches and balanced the number of real and synthetic patches at a 50:50 ratio. 
Additionally, we set aside 25\% of the patches used in training the diffusion model for inference in downstream tasks. 
This approach ensures the inference set includes a sufficient proportion of each nucleus class, enabling a comprehensive comparison of classification results.
%
%
%
We used Hover-Net~\cite{graham2019hover} as a baseline network, a neural network designed to segment clustered nuclei by predicting the horizontal and vertical distances between individual nucleus pixels and their respective centers of mass. 
Since Hover-Net aims to predict horizontal and vertical maps to improve nuclei instance segmentation performance, it requires instance labels to generate ground truth distance masks. 
Therefore, we excluded methods that do not generate distance maps, such as SemanticPalette~\cite{semantic-palette} and Park~\textit{et al.}~\cite{gcdp}.
%
In addition, even though Yu~\textit{et al.}~\cite{diffaug} generates distance maps during unconditional label generation, the label quality was poor, so we excluded it from the comparison. 
%

\input{tables/downstream_seg}

Therefore, we chose to compare our method with SDM~\cite{sdm} (see Table~\ref{tab:downstream}). \
%
%
We conducted experiments with the following configurations: Baseline (with conventional augmentations), SDM, and our method. 
For the dataset used in this task, we generated an equal number of patches using both the SDM method and our approach.
%
%
%
%
Table~\ref{tab:downstream} shows the nuclei segmentation and classification performance using the baseline method.
For segmentation, the Dice coefficient and Aggregated Jaccard Index (AJI) metrics are employed to measure performance for semantic and instance segmentation performance, respectively.
For classification, $F^{c_i}$ represents the $F_1$ score for the $i$-th nulceus class (type) and $F_d$ indicates the detection quality to measure the quality of instance detection.
%
%
%
We analyzed the performance improvements resulting from the application of conventional augmentation (denoted as w/ Aug.), as well as the addition of SDM and our synthetic data.
%
%
In particular, our synthetic data led to significant improvements across all datasets compared to using conventional augmentation alone. In the Lizard dataset, our method secured second place for most metrics by small margins, typically less than 0.5\%.
%
Especially, in the case of AJI metrics which implies instance segmentation performance, there was a gap of around 0.4\%.
Given that SDM relies on complete labeling for conditions and demonstrates high efficiency in generating images matching these label conditions, this suggests that our masks, employed in this downstream task, are effectively created, contributing to robust performance.
In the PanNuke and EndoNuke datasets, our approach predominantly achieved first place, demonstrating robustness across different tissue types and staining modalities in histopathology datasets.

%
%
%
%
%
%

%

%
%

\begin{figure}[t]
  \centering
  \includegraphics[width=\textwidth]{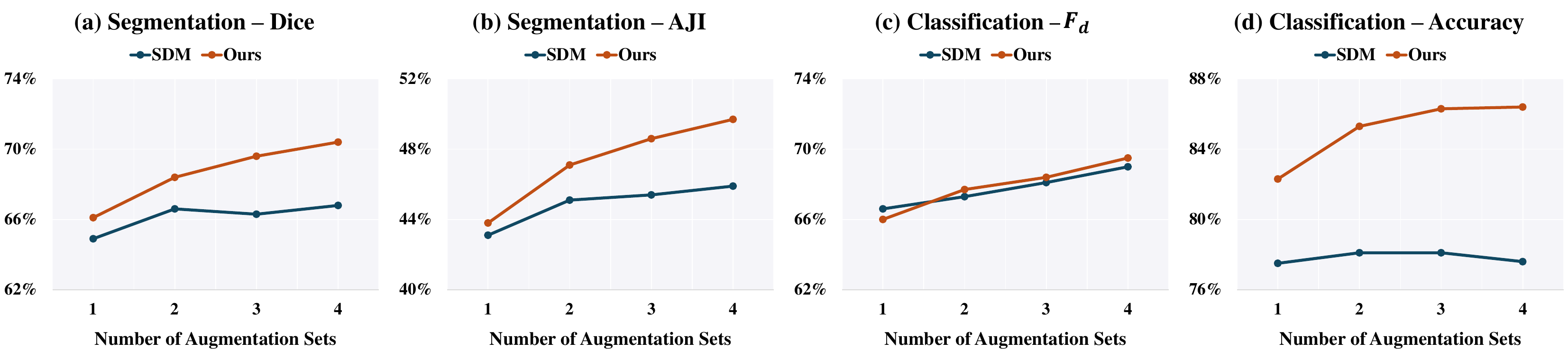} 
  \caption{Benefits of point condition-based diverse label generation on downstream tasks}
  \label{graph:vs_sdm}
\end{figure}

Furthermore, we demonstrate the effectiveness of generating diverse labels by our scheme (see Fig.~\ref{fig:div_layout}) for data augmentation, as shown in Fig.~\ref{graph:vs_sdm}.
%
%
%
In this experiment, we trained the models exclusively with synthetic data by increasing the number of augmentation sets.
%
%
Our approach is compared to SDM, which generates images based on full-pixel labels.
%
%
To perform this comparison, we prepared point layouts for our method and full-pixel labels for SDM.
As we increased the number of synthetic sets, we evaluated the downstream segmentation and classification performance on the Lizard dataset. 
As shown in Fig.~\ref{graph:vs_sdm} (a) and (b), SDM exhibits performance saturation in both the Dice and AJI metrics at a lower point compared to our method, demonstrating continuous improvement.
%
Moreover, as shown in Fig.~\ref{graph:vs_sdm} (c) and (d), our method achieves higher accuracy (an increase by over 10\%) compared to SDM when the number of synthetic sets reaches 4, which also reflects a higher $F_d$ value.
These results indicate that our synthetic data, with its diverse labels, leads to a more diverse set of synthetic image samples, effectively improving the data distribution.

%% file: tables/methods_v1.tex
\begin{table*}[t!]
\centering
\caption{Comparative overview of generative models by conditional inputs and output targets.
}
\renewcommand{\arraystretch}{1.15}
\setlength{\tabcolsep}{2.7pt}
\label{table:methods}
\begin{tabular}{
>{\small}l|
>{\small}l|
>{\small}c
>{\small}c
>{\small}c
}
\toprule
\specialrule{0.1pt}{0pt}{0pt}

\multirow{2}{*}{\textbf{Method}} & \multirow{2}{*}{\textbf{Condition}} & \multicolumn{3}{c}{\textbf{Generation Target}}\\

& & \medium{Image} & \medium{Dist.Map} & \medium{Label}\\
\hline\hline

\medium{Yu~\textit{et al.} \cite{diffaug}} & \medium{None} &
\checkmark  & \checkmark  & \checkmark \\

\medium{SemanticPalette \cite{semantic-palette}} & \medium{Class Proportion} &
\checkmark  & -  & \checkmark \\

\medium{Park~\textit{et al.} \cite{gcdp}} & \medium{Text} &
\checkmark  & -  & \checkmark \\

\medium{SDM \cite{sdm}} & \medium{Label, Instance Edge} &
\checkmark & - & - \\

\hline
\medium{\textbf{Ours}} & \medium{Text, Point Map} &
\checkmark  & \checkmark  & \checkmark \\

\toprule
\specialrule{0.1pt}{0pt}{0pt}
\end{tabular}

\end{table*}

%% file: tables/fid_is_fsd.tex
\begin{table*}[t!]
\centering
\caption{Comparative results of generative models, evaluated by FID, IS, and FSD. The best results are in \textbf{bold} and the second best are \underline{underlined}.}
\renewcommand{\arraystretch}{1.15}
\setlength{\tabcolsep}{2.7pt}
\label{table:fid_is_fsd}
\begin{tabular}{
>{\small}l|
>{\small}c
>{\small}c
>{\small}c|
>{\small}c
>{\small}c
>{\small}c|
>{\small}c
>{\small}c
>{\small}c
}
\toprule
\specialrule{0.1pt}{0pt}{0pt}

\multirow{2}{*}{\textbf{Method}} & \multicolumn{3}{c|}{\textbf{Lizard}} & \multicolumn{3}{c|}{\textbf{PanNuke}} & \multicolumn{3}{c}{\textbf{EndoNuke}} \\
\multicolumn{1}{c|}{} & \medium{FID$\downarrow$} & \medium{IS$\uparrow$} & \medium{FSD$\downarrow$} & \medium{FID$\downarrow$} & \medium{IS$\uparrow$} & \medium{FSD$\downarrow$} & \medium{FID$\downarrow$} & \medium{IS$\uparrow$} & \medium{FSD$\downarrow$} \\
\hline\hline

\medium{Yu~\textit{et al.}~\cite{diffaug}}
& \medium{-} & \medium{-} & \medium{963.36}
& \medium{-} & \medium{-} & \medium{1292.05}
& \medium{-} & \medium{-} & \medium{931.21}
\\

\medium{SemanticPalette~\cite{semantic-palette}} 
& \medium{86.17} & \medium{2.11} & \underline{\medium{0.55}}
& \medium{109.23} & \medium{3.36} & \textbf{\medium{1.23}}
& \medium{90.00} & \medium{1.40} & \textbf{\medium{1.88}}
\\
 
\medium{Park~\textit{et al.}~\cite{gcdp}} 
& \medium{52.65} & \medium{2.22} & \medium{65.06}
& \underline{\medium{61.16}} & \medium{3.48} & \medium{34.43}
& \textbf{\medium{52.99}} & \medium{1.88} & \medium{110.00}
\\

\medium{SDM~\cite{sdm}}
& \underline{\medium{45.99}} & \underline{\medium{2.35}} & \medium{-}
& \medium{107.80} & \textbf{\medium{3.82}} & \medium{-}
& \medium{105.17} & \textbf{\medium{2.27}} & \medium{-}
\\
\hline

\medium{Ours w/o $pc$} 
& \medium{69.10} & \medium{2.02} & \medium{109.18}
& \medium{-} & \medium{-} & \medium{-}
& \medium{-} & \medium{-} & \medium{-}
\\

\textbf{\medium{Ours}}
& \textbf{\medium{38.78}} & \textbf{\medium{2.40}} & \textbf{\medium{0.13}}
& \textbf{\medium{37.35}} & \underline{\medium{3.77}} & \underline{\medium{1.44}}
& \underline{\medium{69.94}} & \underline{\medium{2.17}} & \underline{\medium{29.57}}
\\

\toprule
\specialrule{0.1pt}{0pt}{0pt}
\end{tabular}


\end{table*}

%% file: tables/mdice_v2.tex
\begin{table*}[tb!]
\centering
\caption{
    Comparative results of instance separation algorithms, evaluated by mDice.
    The best results are in \textbf{bold} and the second best are \underline{underlined}.
}
\renewcommand{\arraystretch}{1.15}
\setlength{\tabcolsep}{2.7pt}
\label{table:mdice}
\begin{tabular}{
>{\medium}l|
>{\medium}c
>{\medium}c
>{\medium}c
}
\toprule
\specialrule{0.1pt}{0pt}{0pt}

\multirow{2}{*}{\textbf{\small{Method}}} & \multicolumn{3}{c}{\textbf{\small{mDice}}} \\
\multicolumn{1}{c|}{} & {Lizard} & {PanNuke} & {EndoNuke} \\

\hline\hline

\scriptsize{Connectivity-based} & \underline{{0.9383}} & {0.9146} & {0.5524} \\
 
Yu~\textit{et al.}~\cite{diffaug} & {0.9374} & \underline{{0.9462}} & \underline{{0.9268}} \\
 
\hline

\textbf{Ours} & \textbf{{0.9754}} & \textbf{{0.9980}} & \textbf{{0.9634}} \\

\toprule
\specialrule{0.1pt}{0pt}{0pt}
\end{tabular}

\end{table*}

%% file: tables/downstream_seg.tex


\begin{table}[htb!]
\caption{
Downstream segmentation and classification performance comparison for various augmentation methods using Hover-Net~\cite{graham2019hover} as a baseline network.
The best results are in \textbf{bold} and the second best are \underline{underlined}.
}
\label{tab:downstream}
\centering
\renewcommand{\arraystretch}{1.04}
\renewcommand{\tabcolsep}{3.0pt}
\begin{tabular}{
>{\scriptsize}l|
>{\scriptsize}l|
>{\qhrh}c
>{\qhrh}c|
>{\qhrh}c
>{\qhrh}c
>{\qhrh}c
>{\qhrh}c
>{\qhrh}c
>{\qhrh}c
>{\qhrh}c
>{\qhrh}c
}
\toprule\specialrule{0.05pt}{0pt}{0pt}
\multirow{2}{*}{\textbf{Dataset}} & \multirow{2}{*}{\textbf{Method}} & \multicolumn{2}{c|}{\textbf{\scriptsize{Segmentation}}} & \multicolumn{8}{c}{\textbf{\scriptsize{Classification}}} \\ \cline{3-12}
\multicolumn{1}{c|}{} & \multicolumn{1}{c|}{} & Dice & AJI & $F_d$ & Acc & $F^{{c_1}}$ & $F^{c_2}$ & $F^{c_3}$ & $F^{c_4}$ & $F^{c_5}$ & $F^{c_6}$
\\ 
\hline\hline
 \multirow{4}{*}{\textbf{Lizard}} 
& Baseline & 0.620 & 0.383 & 0.619 & 0.763 & 0.012 & 0.548 & 0.318 & 0.146 & 0.050 & 0.252 \\
& w/ Aug. & 0.676 & 0.425 & 0.646 & 0.818 & 0.062 & 0.599 & 0.351 & 0.268 & 0.264 & 0.367 \\ 
& w/ SDM  & \textbf{0.718} & \textbf{0.488} & \textbf{0.699} & \underline{0.862} & \textbf{0.185} & \textbf{0.679} & \underline{0.413} & \textbf{0.350} & \underline{0.333} & \textbf{0.455} \\
& w/ \textbf{Ours} & \underline{0.716} & \underline{0.484} & \underline{0.694} & \textbf{0.866} & \underline{0.161} & \underline{0.676} & \textbf{0.434} & \underline{0.346} & \textbf{0.341} & \underline{0.447} \\

\hline
 \multirow{4}{*}{\textbf{PanNuke}} 
& Baseline    & 0.782 & 0.598 & 0.763 & 0.668 & 0.420 & 0.356 & 0.102 & 0.301 & 0.475 & - \\
& w/ Aug. & 0.816 & 0.641 & 0.791 & \underline{0.708} & \underline{0.492} & 0.394 & 0.107 & \underline{0.380} & 0.524 & - \\ 
& w/ SDM  & \underline{0.821} & \underline{0.654} & \underline{0.800} & 0.702 & 0.481 & \underline{0.398} & \textbf{0.130} & 0.336 & \underline{0.528} & - \\
& w/ \textbf{Ours} & \textbf{0.824} & \textbf{0.662} & \textbf{0.806} & \textbf{0.736} & \textbf{0.516} & \textbf{0.434} & \underline{0.127} & \textbf{0.420} & \textbf{0.561} & - \\

\hline
 \multirow{4}{*}{\textbf{EndoNuke}} 
& Baseline    & 0.878 & 0.594 & 0.815 & 0.891 & 0.734 & 0.504 & 0.013 & - & - & - \\
& w/ Aug. & 0.889 & 0.602 & 0.820 & 0.905 & 0.747 & 0.598 & 0.008 & - & - & - \\ 
& w/ SDM  & \textbf{0.900} & \underline{0.642} & \textbf{0.848} & \underline{0.909} & \underline{0.768} & \underline{0.654} & 0.013 & - & - & - \\
& w/ \textbf{Ours} & \underline{0.899} & \textbf{0.645} & \underline{0.844} & \textbf{0.926} & \textbf{0.787} & \textbf{0.665} & 0.008 & - & - & - \\

\toprule\specialrule{0.05pt}{0pt}{0pt}
\end{tabular}

\end{table}

%% file: 05_conclusion.tex
\section{Conclusion}
\label{sec:conclusion}

In this work, we introduced a novel approach to concurrently generate image-label pairs for histopathology nuclei images. 
We model the joint distribution of image, semantic label, and distance map using a single joint diffusion model. 
In addition, we introduced two context conditioning methods, a point map and text prompts, 
to enable precise control over the label synthesis process and faithful synthesis of histopathology images. 
Lastly, we use the synthesized distance mask to obtain instance label maps which are useful for downstream tasks such as nuclei instance segmentation. 

For future work, we plan to reduce the time cost for data synthesis while maintaining the quality of the sampling to address the challenges of time-efficient synthesized data collection. 
Although we can employ existing methods (e.g., Abousamra~\textit{et al}.~\cite{abousamra2023topology}) to generate the input point layout, developing a generative method for synthesizing more realistic point layouts is another research direction to explore.


%% file: 06_supplementary.tex
\begin{center}
    \addcontentsline{toc}{chapter}{Appendix}
    \LARGE \textbf{Appendix}
\end{center}

In this supplementary material, we provide details on the pathologists' assessment and more qualitative results on PanNuke~\cite{pannuke}, Lizard~\cite{lizard}, and EndoNuke~\cite{endonuke}.

\section*{Appendix A. Qualitative Assessment by Pathologists}

\begin{figure}[]
    \centering
    \includegraphics[width=10.0cm, keepaspectratio]{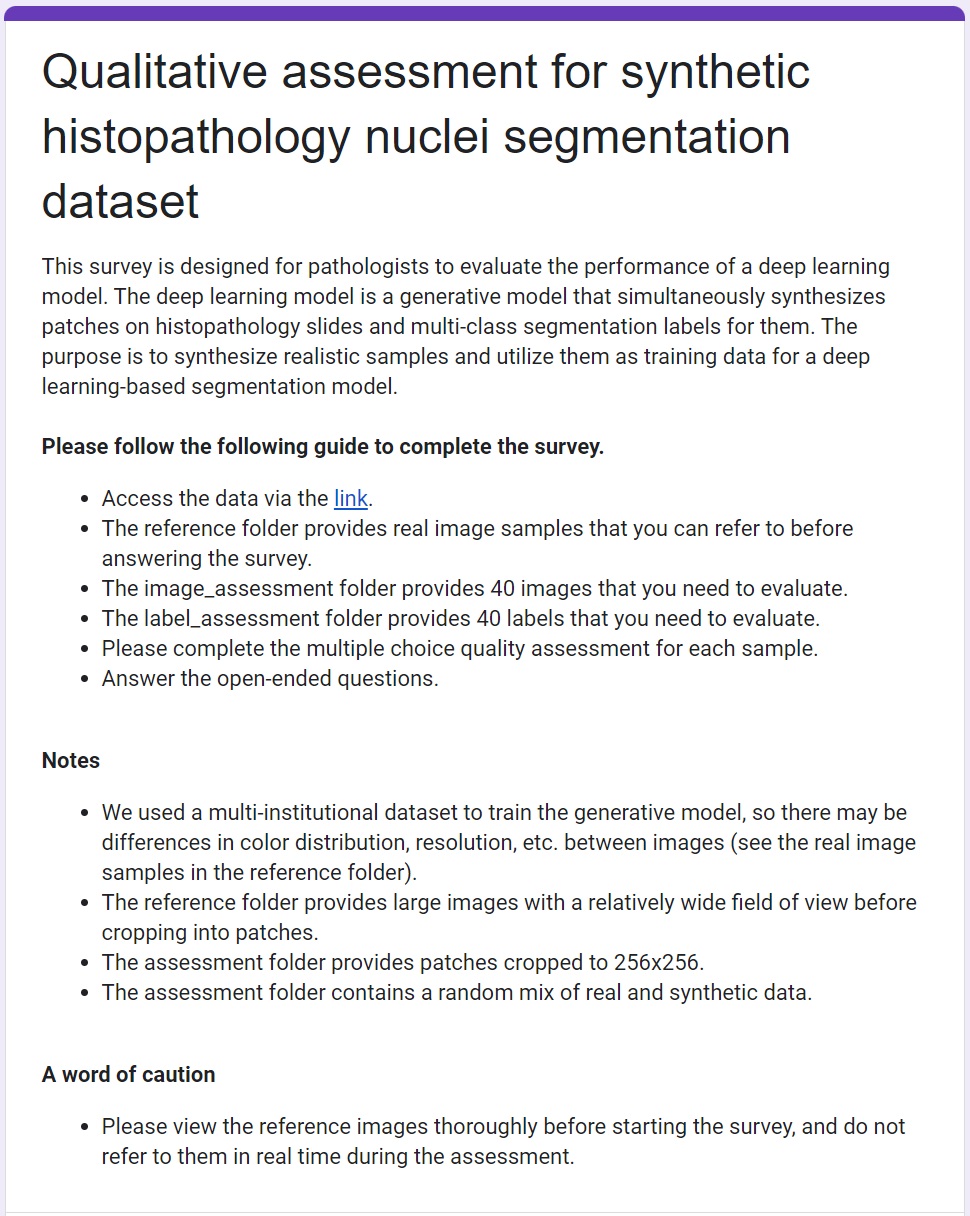}
    \caption{
    Introduction to the qualitative assessment of synthetic data survey.
    }
    \label{fig:survey}
\end{figure}

Fig.~\ref{fig:survey} shows the instructions for the qualitative assessment survey conducted by pathologists. The following text describes the open-response questions asked in the survey and the pathologists' responses.

\noindent\textbf{Q1. What were the key considerations when evaluating image quality?}
\begin{itemize}
    \item Whether the image represent a tissue morphology that could exist in reality.
    \item Whether nuclei membranes, nucleoli, and chromatin can be distinguished.
    \item Whether the boundaries between cells can be seen, including resolution, noise, and color contrast.
    \item Good match to actual histologic structures
    \item Diagnostic potential of the tissue.
\end{itemize}

\noindent\textbf{Q2. What were the key considerations when evaluating label quality?}
\begin{itemize}
    \item Errors seem to occur when connective tissue cells are in between inflammation, but evaluation of this is limited due to difficulty in accurately determining GT for these cells.
    \item Differentiation of epithelial and inflammatory cell.
    \item Eosinophil matching, differentiation of (1) neutrophil and karyorrhexis, (2) plasma and stromal cell, (3) lymphocyte and degenerated epithelial cell.
    \item Whether to differentiate between lymphocytes and connective tissue, which are relatively difficult to distinguish.
    \item Identify the location and type of cell.
\end{itemize}

\noindent\textbf{Q3. What could be improved in the synthetic images?}
\begin{itemize}
    \item Reproduction of polarity loss and disorientation depending on the actual malignancy of the cell.
    \item Sharpness (resolution), finer differentiation of nuclei.
    \item H\&E stain is too intense or too light in some areas.
    \item Some blurry or fragmented images.
\end{itemize}

\noindent\textbf{Q4. What could be improved in the synthetic labels?}
\begin{itemize}
    \item Error occurs when connective tissue is between inflammation.
    \item Reliable differentiation of epithelial cells.
    \item Differentiation of (1) neutrophil and karyorrhexis, (2) lymphocytes and fibroblasts in connective tissue, and (3) connective tissue and lymphocyte.
\end{itemize}

\noindent\textbf{Q5. Please feel free to provide any additional comments.}
\begin{itemize}
    \item Since there were only three possible choices to the question, I graded more conservatively, but overall, the synthetic image quality and label performance were good.
\end{itemize}

\section*{Appendix B. Color Quality Evaluation}
\label{sec:color}

\begin{figure}[]
    \centering
    \includegraphics[width=\textwidth, keepaspectratio]{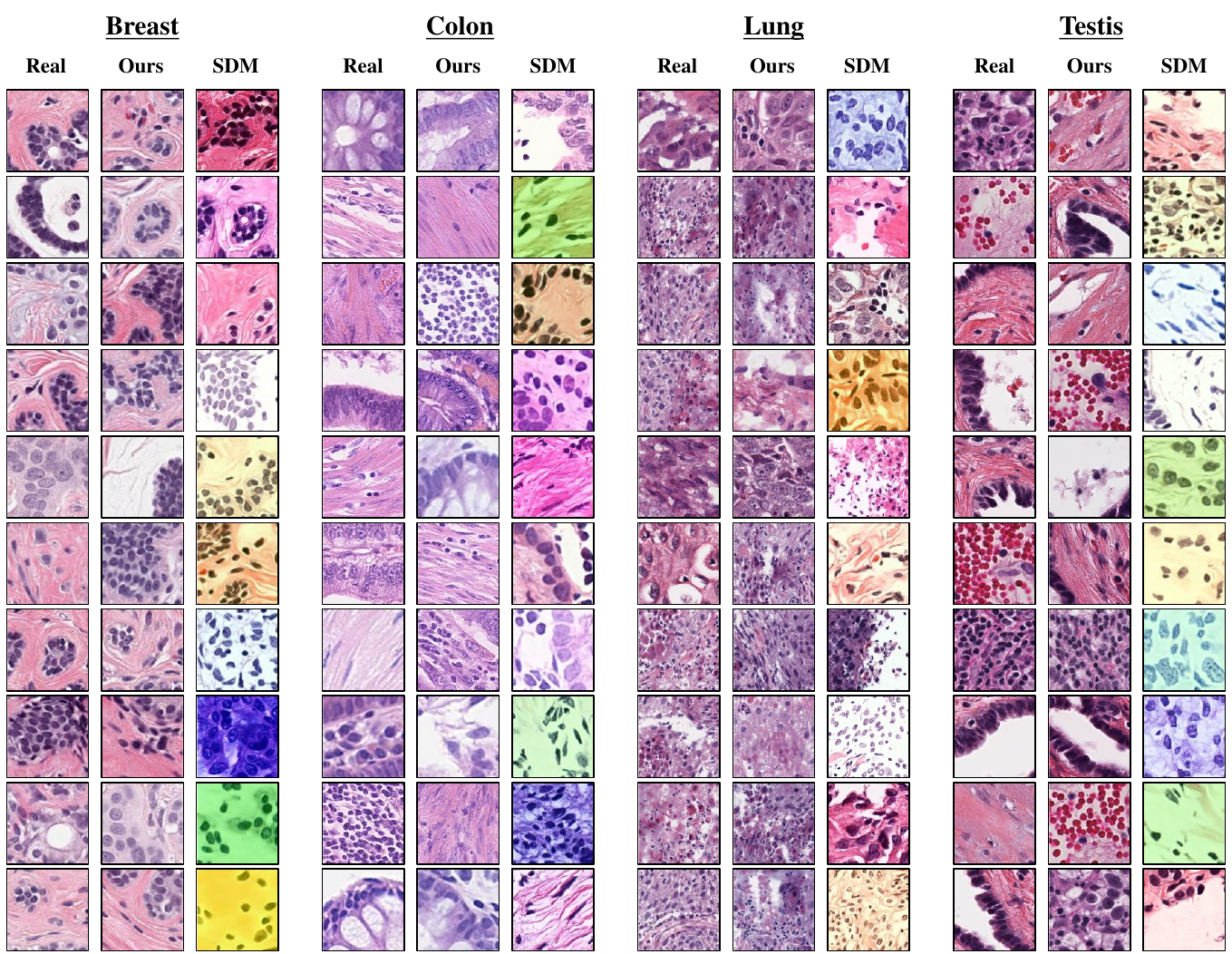}
    \caption{
    Qualitative result on PanNuke.
    }
    \label{aq_pan}
\end{figure}
\begin{figure}[]
    \centering
    \includegraphics[width=\textwidth, keepaspectratio]{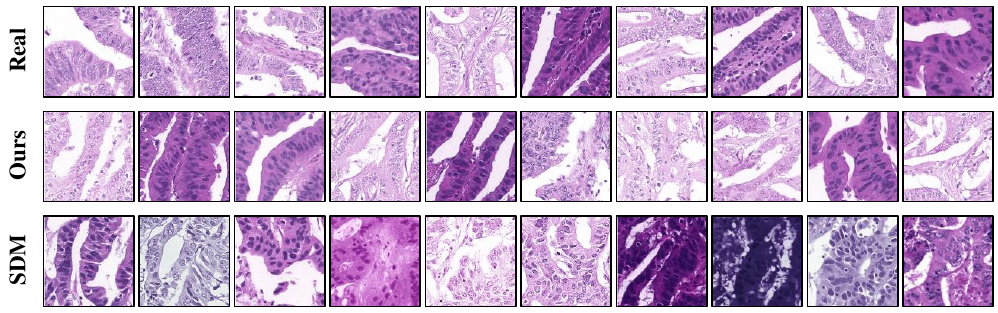}
    \caption{
    Qualitative result on Lizard.
    }
    \label{aq_liz}
\end{figure}
\begin{figure}[]
    \centering
    \includegraphics[width=\textwidth, keepaspectratio]{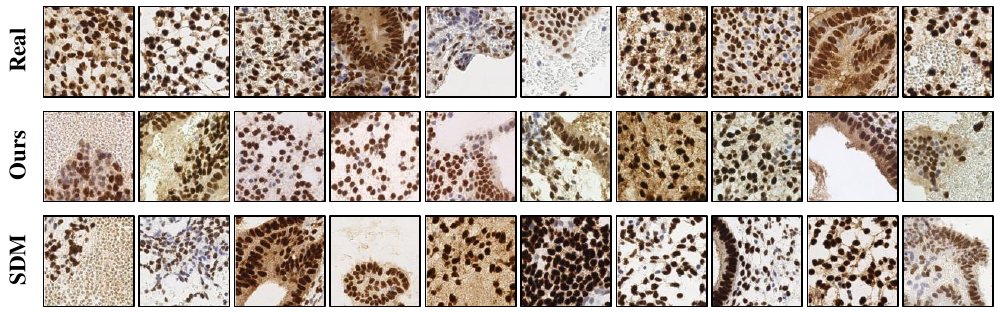}
    \caption{
    Qualitative result on EndoNuke.
    }
    \label{aq_endo}
\end{figure}

To evaluate the color quality, we present several synthetic and real images for each dataset as depicted in Figs.~\ref{aq_pan}-\ref{aq_endo}.
Our method consistently mimics the color distribution of the real data.
Moreover, Fig.~\ref{aq_pan} shows that our method generates images considering the color features that vary depending on the tissue type.
SDM-generated samples exhibit unrealistic but diverse colors, resulting in a high Inception Score (IS).

%% file: main.bbl
\begin{thebibliography}{10}
\providecommand{\url}[1]{\texttt{#1}}
\providecommand{\urlprefix}{URL }
\providecommand{\doi}[1]{https://doi.org/#1}

\bibitem{abousamra2023topology}
Abousamra, S., Gupta, R., Kurc, T., Samaras, D., Saltz, J., Chen, C.: Topology-guided multi-class cell context generation for digital pathology. In: Proceedings of the IEEE/CVF Conference on Computer Vision and Pattern Recognition. pp. 3323--3333 (2023)

\bibitem{sb_gan}
Azadi, S., Tschannen, M., Tzeng, E., Gelly, S., Darrell, T., Lucic, M.: Semantic bottleneck scene generation. arXiv preprint arXiv:1911.11357  (2019)

\bibitem{dataset-ddpm}
Baranchuk, D., Rubachev, I., Voynov, A., Khrulkov, V., Babenko, A.: Label-efficient semantic segmentation with diffusion models. arXiv preprint arXiv:2112.03126  (2021)

\bibitem{FSD}
Bau, D., Zhu, J.Y., Wulff, J., Peebles, W., Strobelt, H., Zhou, B., Torralba, A.: Seeing what a gan cannot generate. In: Proceedings of the IEEE/CVF International Conference on Computer Vision. pp. 4502--4511 (2019)

\bibitem{sharp-gan}
Butte, S., Wang, H., Xian, M., Vakanski, A.: Sharp-gan: Sharpness loss regularized gan for histopathology image synthesis. In: 2022 IEEE 19th International Symposium on Biomedical Imaging (ISBI). pp.~1--5. IEEE (2022)

\bibitem{cheng2020computational}
Cheng, J., Han, Z., Mehra, R., Shao, W., Cheng, M., Feng, Q., Ni, D., Huang, K., Cheng, L., Zhang, J.: Computational analysis of pathological images enables a better diagnosis of tfe3 xp11. 2 translocation renal cell carcinoma. Nature communications  \textbf{11}(1), ~1778 (2020)

\bibitem{diffusion_beats_gan}
Dhariwal, P., Nichol, A.: Diffusion models beat gans on image synthesis. Advances in neural information processing systems  \textbf{34},  8780--8794 (2021)

\bibitem{doan2022sonnet}
Doan, T.N., Song, B., Vuong, T.T., Kim, K., Kwak, J.T.: Sonnet: A self-guided ordinal regression neural network for segmentation and classification of nuclei in large-scale multi-tissue histology images. IEEE Journal of Biomedical and Health Informatics  \textbf{26}(7),  3218--3228 (2022)

\bibitem{pannuke}
Gamper, J., Koohbanani, N.A., Benes, K., Graham, S., Jahanifar, M., Khurram, S.A., Azam, A., Hewitt, K., Rajpoot, N.: Pannuke dataset extension, insights and baselines. arXiv preprint arXiv:2003.10778  (2020)

\bibitem{gong2021style}
Gong, X., Chen, S., Zhang, B., Doermann, D.: Style consistent image generation for nuclei instance segmentation. In: Proceedings of the IEEE/CVF winter conference on applications of computer vision. pp. 3994--4003 (2021)

\bibitem{gan}
Goodfellow, I., Pouget-Abadie, J., Mirza, M., Xu, B., Warde-Farley, D., Ozair, S., Courville, A., Bengio, Y.: Generative adversarial nets. Advances in neural information processing systems  \textbf{27} (2014)

\bibitem{lizard}
Graham, S., Jahanifar, M., Azam, A., Nimir, M., Tsang, Y.W., Dodd, K., Hero, E., Sahota, H., Tank, A., Benes, K., et~al.: Lizard: A large-scale dataset for colonic nuclear instance segmentation and classification. In: Proceedings of the IEEE/CVF International Conference on Computer Vision. pp. 684--693 (2021)

\bibitem{graham2019hover}
Graham, S., Vu, Q.D., Raza, S.E.A., Azam, A., Tsang, Y.W., Kwak, J.T., Rajpoot, N.: Hover-net: Simultaneous segmentation and classification of nuclei in multi-tissue histology images. Medical image analysis  \textbf{58},  101563 (2019)

\bibitem{transnuseg}
He, Z., Unberath, M., Ke, J., Shen, Y.: Transnuseg: A lightweight multi-task transformer for nuclei segmentation. In: International Conference on Medical Image Computing and Computer-Assisted Intervention. pp. 206--215. Springer (2023)

\bibitem{FID}
Heusel, M., Ramsauer, H., Unterthiner, T., Nessler, B., Hochreiter, S.: Gans trained by a two time-scale update rule converge to a local nash equilibrium. Advances in neural information processing systems  \textbf{30} (2017)

\bibitem{dpm}
Ho, J., Jain, A., Abbeel, P.: Denoising diffusion probabilistic models. Advances in neural information processing systems  \textbf{33},  6840--6851 (2020)

\bibitem{hoogeboom}
Hoogeboom, E., Nielsen, D., Jaini, P., Forr{\'e}, P., Welling, M.: Argmax flows and multinomial diffusion: Learning categorical distributions. Advances in Neural Information Processing Systems  \textbf{34},  12454--12465 (2021)

\bibitem{hou2019robust}
Hou, L., Agarwal, A., Samaras, D., Kurc, T.M., Gupta, R.R., Saltz, J.H.: Robust histopathology image analysis: To label or to synthesize? In: Proceedings of the IEEE/CVF Conference on Computer Vision and Pattern Recognition. pp. 8533--8542 (2019)

\bibitem{semantic-palette}
Le~Moing, G., Vu, T.H., Jain, H., P{\'e}rez, P., Cord, M.: Semantic palette: Guiding scene generation with class proportions. In: Proceedings of the IEEE/CVF Conference on Computer Vision and Pattern Recognition. pp. 9342--9350 (2021)

\bibitem{insmix}
Lin, Y., Wang, Z., Cheng, K.T., Chen, H.: Insmix: Towards realistic generative data augmentation for nuclei instance segmentation. In: International Conference on Medical Image Computing and Computer-Assisted Intervention. pp. 140--149. Springer (2022)

\bibitem{diffuse_morph}
Moghadam, P.A., Van~Dalen, S., Martin, K.C., Lennerz, J., Yip, S., Farahani, H., Bashashati, A.: A morphology focused diffusion probabilistic model for synthesis of histopathology images. In: Proceedings of the IEEE/CVF Winter Conference on Applications of Computer Vision. pp. 2000--2009 (2023)

\bibitem{endonuke}
Naumov, A., Ushakov, E., Ivanov, A., Midiber, K., Khovanskaya, T., Konyukova, A., Vishnyakova, P., Nora, S., Mikhaleva, L., Fatkhudinov, T., et~al.: Endonuke: Nuclei detection dataset for estrogen and progesterone stained ihc endometrium scans. Data  \textbf{7}(6), ~75 (2022)

\bibitem{dataset-diffusion}
Nguyen, Q., Vu, T., Tran, A., Nguyen, K.: Dataset diffusion: Diffusion-based synthetic dataset generation for pixel-level semantic segmentation. arXiv preprint arXiv:2309.14303  (2023)

\bibitem{nishimura2019weakly}
Nishimura, K., Ker, D.F.E., Bise, R.: Weakly supervised cell instance segmentation by propagating from detection response. In: Medical Image Computing and Computer Assisted Intervention--MICCAI 2019: 22nd International Conference, Shenzhen, China, October 13--17, 2019, Proceedings, Part I 22. pp. 649--657. Springer (2019)

\bibitem{diffmix}
Oh, H.J., Jeong, W.K.: Diffmix: Diffusion model-based data synthesis for nuclei segmentation and classification in imbalanced pathology image datasets. In: Medical Image Computing and Computer Assisted Intervention -- MICCAI 2023. pp. 337--345. Springer (2023)

\bibitem{gcdp}
Park, M., Yun, J., Choi, S., Choo, J.: Learning to generate semantic layouts for higher text-image correspondence in text-to-image synthesis. In: Proceedings of the IEEE/CVF International Conference on Computer Vision. pp. 7591--7600 (2023)

\bibitem{IS}
Salimans, T., Goodfellow, I., Zaremba, W., Cheung, V., Radford, A., Chen, X.: Improved techniques for training gans. Advances in neural information processing systems  \textbf{29} (2016)

\bibitem{nasdm}
Shrivastava, A., Fletcher, P.T.: Nasdm: Nuclei-aware semantic histopathology image generation using diffusion models. In: Medical Image Computing and Computer Assisted Intervention -- MICCAI 2023. pp. 786--796. Springer (2023)

\bibitem{vahadane}
Vahadane, A., Peng, T., Sethi, A., Albarqouni, S., Wang, L., Baust, M., Steiger, K., Schlitter, A.M., Esposito, I., Navab, N.: Structure-preserving color normalization and sparse stain separation for histological images. IEEE transactions on medical imaging  \textbf{35}(8),  1962--1971 (2016)

\bibitem{verghese2023computational}
Verghese, G., Lennerz, J.K., Ruta, D., Ng, W., Thavaraj, S., Siziopikou, K.P., Naidoo, T., Rane, S., Salgado, R., Pinder, S.E., et~al.: Computational pathology in cancer diagnosis, prognosis, and prediction--present day and prospects. The Journal of Pathology  \textbf{260}(5),  551--563 (2023)

\bibitem{wtsd}
Van~der Walt, S., Sch{\"o}nberger, J.L., Nunez-Iglesias, J., Boulogne, F., Warner, J.D., Yager, N., Gouillart, E., Yu, T.: scikit-image: image processing in python. PeerJ  \textbf{2}, ~e453 (2014)

\bibitem{sian}
Wang, H., Xian, M., Vakanski, A., Shareef, B.: Sian: style-guided instance-adaptive normalization for multi-organ histopathology image synthesis. In: 2023 IEEE 20th International Symposium on Biomedical Imaging (ISBI). pp.~1--5. IEEE (2023)

\bibitem{sdm}
Wang, W., Bao, J., Zhou, W., Chen, D., Chen, D., Yuan, L., Li, H.: Semantic image synthesis via diffusion models. arXiv preprint arXiv:2207.00050  (2022)

\bibitem{rrdb}
Wang, X., Yu, K., Wu, S., Gu, J., Liu, Y., Dong, C., Qiao, Y., Change~Loy, C.: Esrgan: Enhanced super-resolution generative adversarial networks. In: Proceedings of the European conference on computer vision (ECCV) workshops. pp.~0--0 (2018)

\bibitem{watershed}
Yang, X., Li, H., Zhou, X.: Nuclei segmentation using marker-controlled watershed, tracking using mean-shift, and kalman filter in time-lapse microscopy. IEEE Transactions on Circuits and Systems I: Regular Papers  \textbf{53}(11),  2405--2414 (2006)

\bibitem{attribute_gan}
Ye, J., Xue, Y., Liu, P., Zaino, R., Cheng, K.C., Huang, X.: A multi-attribute controllable generative model for histopathology image synthesis. In: Medical Image Computing and Computer Assisted Intervention--MICCAI 2021: 24th International Conference, Strasbourg, France, September 27--October 1, 2021, Proceedings, Part VIII 24. pp. 613--623. Springer (2021)

\bibitem{pathldm2024}
Yellapragada, S., Graikos, A., Prasanna, P., Kurc, T., Saltz, J., Samaras, D.: Pathldm: Text conditioned latent diffusion model for histopathology. In: Proceedings of the IEEE/CVF Winter Conference on Applications of Computer Vision. pp. 5182--5191 (2024)

\bibitem{diffaug}
Yu, X., Li, G., Lou, W., Liu, S., Wan, X., Chen, Y., Li, H.: Diffusion-based data augmentation for nuclei image segmentation. In: International Conference on Medical Image Computing and Computer-Assisted Intervention. pp. 592--602. Springer (2023)

\bibitem{dataset-gan}
Zhang, Y., Ling, H., Gao, J., Yin, K., Lafleche, J.F., Barriuso, A., Torralba, A., Fidler, S.: Datasetgan: Efficient labeled data factory with minimal human effort. In: Proceedings of the IEEE/CVF Conference on Computer Vision and Pattern Recognition. pp. 10145--10155 (2021)

\end{thebibliography}
